# On the Average Similarity Degree between Solutions of Random *k*-SAT and Random CSPs[*]

Ke Xu  and  Wei Li

*National Laboratory of Software Development Environment*
*Department of Computer Science and Engineering*
*Beijing University of Aeronautics and Astronautics, Beijing  100083, P.R. China*

Email[#]: {kexu,liwei}@nlsde.buaa.edu.cn

**Abstract**

To study the structure of solutions for random *k*-SAT and random CSPs, this paper introduces the concept of *average similarity degree* to characterize how solutions are similar to each other. It is proved that under certain conditions, as $r$ (i.e. the ratio of constraints to variables) increases, the limit of average similarity degree when the number of variables approaches infinity exhibits phase transitions at a threshold point, shifting from a smaller value to a larger value abruptly. For random *k*-SAT this phenomenon will occur when $k \geq 5$. It is further shown that this threshold point is also a singular point with respect to $r$ in the asymptotic estimate of the second moment of the number of solutions. Finally, we discuss how this work is helpful to understand the hardness of solving random instances and a possible application of it to the design of search algorithms.

## 1. Introduction

Over the last ten years, following the seminal paper of Cheeseman, Kanefsky and Taylor[6], one of the most exciting areas in artificial intelligence and computer science has been the study of phase transition behaviour in hard combinatorial problems. A lot of experimental and theoretical studies indicate that many problems of practical importance can be characterized by a control parameter in such a way that the space of problem instances is divided into two regions: the under-constrained region where almost all problems have many solutions, and the over-constrained region where almost all problems have no solution, with a sharp transition between them. For example, in the well-studied random 3-SAT, it has been observed empirically that the satisfiability phase transition occurs when the ratio of clauses to variables is approximately 4.3[13]. Similar phenomena were also found for other values of *k* in random *k*-SAT. Up to now, only the phase transition point for 2-SAT has been proved to be 1 by Chvátal and Reed[5] and Goerdt[9]. For random 3-SAT, the best known lower bound and upper bound for the phase transition point are 3.145[1] and 4.602[11] respectively. The interest in the phase transition behaviour has been furthered enhanced by the observation that the instances in the transition region are the hardest to solve. Because of the extreme hardness of the instances at the transition region, such instances provide a useful benchmark for evaluating search algorithms.

---

[*] Research supported by National 973 Project of China Grant No.G1999032701.

[#] For the first author, there is a permanent email address that is ke.xu@263.net.

In addition to the theoretical and experimental interest, the phase transition research is also helpful to understand what makes NP-complete problems so hard to solve and thus hopefully improve the efficiency of algorithms. Starting from the point that the nature of many algorithms is to search solutions in the space of assignments, one can easily find that a good understanding of the phase transition behaviour will undoubtedly require a deep understanding of the structure of solutions, e.g. how solutions are distributed in the space of assignments. In recent years, there have been studies about the structure of solutions in the SAT phase transition. Parkes showed experimentally that a significant subclass of instances emerges when crossing the satisfiability phase transition[16]. In such instances, the solutions are not randomly distributed but all lie in a cluster that is exponentially large. By means of the replica method from statistic mechanics, Monasson et al. used the study of how variables freeze to a single value to investigate the transition from P to NP[14]. Similarly, Biroli et al. studied the typical Hamming distance between two solutions of a random 3-SAT problem[4]. It should be noted that the validity of the replica method depends on a set of unproven assumptions that are not generally believed to be true[3,17]. From a theoretical point of view, it is therefore very essential to study the structure of solutions by use of mathematical (rigorous) methods.

In fact, SAT is a special case of the *constraint satisfaction problem* (CSP) which is defined as follows.

**Definition 1.1** *A constraint satisfaction problem*

A *constraint satisfaction problem* consists of a finite set $U = \{u_1, \cdots, u_n\}$ of $n$ variables and a set of constraints. For each variable $u_i$ a *domain* $D_i$ with $d_i$ elements is specified; a variable can only be assigned a value from its domain. For $2 \leq k \leq n$ a *constraint* $C_{i1,i2,\cdots,ik}$ consists of a subset $\{u_{i1}, u_{i2}, \cdots, u_{ik}\}$ of $U$ and a relation $R_{i1,i2,\cdots,ik} \subseteq D_{i1} \times \cdots \times D_{ik}$, where $i1, i2, \cdots, ik$ are distinct. $C_{i1,i2,\cdots,ik}$ is called a $k$-*ary* constraint which bounds the variables $u_{i1}, \cdots, u_{ik}$. $R_{i1,i2,\cdots,ik}$ specifies all the allowed tuples of values for the variables $u_{i1}, \cdots, u_{ik}$ which are compatible with each other. A *solution* to a CSP is an assignment of a value to each variable from its domain such that all the constraints are satisfied. A constraint $C_{i1,i2,\cdots,ik}$ is satisfied if the tuple of values assigned to the variables $u_{i1}, \cdots, u_{ik}$ is in the relation $R_{i1,i2,\cdots,ik}$. A CSP that has a solution is called *soluble*; otherwise it is called *insoluble*.

CSP is a fundamental problem in artificial intelligence, with numerous applications ranging from vision, language comprehension to scheduling and diagnosis[7]. In general, CSP is NP-complete. Recently, there has been a great amount of interest in the phase transition behaviour of random CSPs, both from an experimental and a theoretical point of view[2,8,18,20,21,23,24]. However, there is still some lack of studies about the structure of solutions of random CSPs. To study the phase transition behaviour of random CSPs, we need first a random CSP model to generate random instances. Standard Model B[8,20], the most commonly used CSP model in previous studies, is only a binary model. To include the well-studied random $k$-SAT in the CSP model, we use the following model in this paper, which is essentially a generalization of standard Model B to the $k$-ary case.

**Model GB**

Step 1. *We select with repetition* $t = rn$ *random constraints. A random constraint is formed by selecting without repetition* $k \geq 2$ *of* $n$ *variables.*

Step 2. *For each constraint we uniformly select without repetition $q$ incompatible tuples of values.*

The parameter $r$ determines how many constraints are in a CSP instance, while $q$ determines how restrictive the constraints are. Assume that in Model GB all the variable domains contain the same number of values $d \geq 2$ and $q$ satisfies $q < d^{k-1}$. Recently, Achlioptas et al.[2] shows that for standard Model B, i.e. the binary case of Model GB, if $q \geq d$, almost all the instances generated following Model B are trivially insoluble as the number of variables approaches infinity. Following the same lines as their proof for Model B[1], we can easily show that if $q \geq d^{k-1}$, Model GB will also suffer from the trivial asymptotic insolubility. In such a case, asymptotically, no solution exists to the generated instances. So in this paper, we will only consider the case of $q < d^{k-1}$. The definition of Model GB can also be found in [24] which investigated the average number of nodes used by the backtracking algorithm on Model GB in the case of $q < d$, and proved that in this case Model GB exhibits non-trivial asymptotic behavior (not trivially soluble or insoluble). It is easy to see that random $k$-SAT is also a special case of Model GB if we set $d$ to 2 and $q$ to 1 respectively. Therefore, the results about Model GB in this paper are also applicable to random $k$-SAT. To study the structure of solutions of random instances generated following Model GB, we will first combine assignments into *assignment pairs*, and then introduce the concepts of *similarity number* and *similarity degree* to describe how the two assignments in an assignment pair are similar to each other. Based on these definitions, the concept of *average similarity degree*, measuring how satisfying assignments (i.e. solutions) are similar to each other, will be introduced. Finally, we will discuss the behaviour of the *average similarity degree* for Model GB as $r$ varies.

**Definition 1.2** *An assignment pair*

An *assignment pair* is an ordered pair $<t_i, t_j>$ of assignments to the variables in $U$, where $t_i = (a_{i1}, a_{i2}, \cdots, a_{in})$ and $t_j = (a_{j1}, a_{j2}, \cdots, a_{jn})$ with $a_{il}, a_{jl} \in D_l$. The set that consists of all the assignment pairs is denoted by $A_{pair}$. An assignment pair $<t_i, t_j>$ satisfies a CSP if and only if both $t_i$ and $t_j$ satisfy this CSP. That is to say, an assignment pair $<t_i, t_j>$ is called a satisfying assignment pair of a CSP if and only if both $t_i$ and $t_j$ are solutions of this CSP.

It is easy to see from the above definition that if we know all the solutions of a CSP, then all the satisfying assignment pairs of this CSP can be formed by combining the solutions into ordered pairs of solutions.

**Definition 1.3** *Similarity number* $S^f : A_{pair} \to \{0,1,2,3,...\}$,

$$S^f(<t_i, t_j>) = \sum_{l=1}^{n} Sam(a_{il}, a_{jl}), \quad (1.1)$$

where the function *Sam* is defined as follows:

$$Sam(a_{il}, a_{jl}) = \begin{cases} 1, & \text{if } a_{il} = a_{jl}, \\ 0, & \text{if } a_{il} \neq a_{jl}. \end{cases} \quad (1.2)$$

The *similarity number* of an assignment pair is equal to the number of variables at which the two assignments of this assignment pair take the identical values. By Definition 1.3 it is easy to see that

$0 \leq S^f(<t_i, t_j>) \leq n$.

**Definition 1.4** *Similarity degree* $s^f : A_{pair} \to R$,

$$s^f(<t_i, t_j>) = \frac{S^f(<t_i, t_j>)}{n}.\quad (1.3)$$

The *similarity degree* of an assignment pair is a measure of how the two assignments in this assignment pair are similar to each other, i.e. the ratio of the *similarity number* to the total number of variables. The larger the value of $s^f(<t_i, t_j>)$, the more similar are the two assignments $t_i$ and $t_j$. By Definition 1.4, it is obvious that $0 \leq s^f(<t_i, t_j>) \leq 1$.

Let $A_s$ be the set of assignment pairs whose *similarity degree* is equal to $s$. A random CSP instance, generated following Model GB, is denoted by $\phi$. Let $A_s^{Sat}$ be the set of assignment pairs that are in $A_s$ and satisfy $\phi$. It is obvious that the cardinality $|A_s^{Sat}|$ is a random variable. The expected value of this variable is denoted by $E(|A_s^{Sat}|)$, i.e. $E(|A_s^{Sat}|)$ stands for the expected number of satisfying assignment pairs whose *similarity degree* is equal to $s$. Given $r$, if $E(|A_{s_1}^{Sat}|) > E(|A_{s_2}^{Sat}|)$ holds for $0 \leq s_1, s_2 \leq 1$, then we can say that $s_1$ plays a more important role in satisfying assignment pairs than $s_2$. Using $E(|A_s^{Sat}|)$ as the weighting factor, the *average similarity degree* of satisfying assignment pairs, denoted by $s_{av}$, is defined as follows.

**Definition 1.5** *Average Similarity degree:*

$$s_{av} = \frac{\sum_{s=\frac{0}{n},\frac{1}{n},\cdots,\frac{n}{n}} s E|A_s^{Sat}|}{\sum_{s=\frac{0}{n},\frac{1}{n},\cdots,\frac{n}{n}} E|A_s^{Sat}|}.\quad (1.4)$$

Recall that satisfying assignment pairs are ordered pairs of solutions. The average similarity degree of satisfying assignment pairs can therefore be regarded as a characteristic of the structure of solutions, measuring how solutions are similar to each other. It is proved that as $r$ varies, the limit $\lim_{n \to \infty} s_{av}$, denoted by $s_{av,\infty}$, suddenly shift from a smaller value to a larger value when $r$ crosses a threshold point. Strictly speaking, we have the following two theorems.

**Theorem 1.1** *Given $k$, $q$, there exists $d_N > 0$ such that for every $d > d_N$, there exists a threshold point $r_{cr}$ (the value of $r_{cr}$ depends on $k$, $q$ and $d$) such that $s_{av,\infty} = s_1(r) < s_1(r_{cr})$ when $0 < r < r_{cr}$, while $s_{av,\infty} = s_3(r) > s_3(r_{cr})$ when $r > r_{cr}$, where $s_1(r)$ and $s_3(r)$, satisfying $s_1(r_{cr}) < s_3(r_{cr})$, are two continuous and strictly increasing functions that depend on $k$, $q$ and $d$.*

**Theorem 1.2** *Given $d$, $q$, there exists $k_N > 0$ such that for every $k > k_N$, there exists a threshold point $r_{cr}$ (the value of $r_{cr}$ depends on $k$, $q$ and $d$) such that $s_{av,\infty} = s_1(r) < s_1(r_{cr})$ when $0 < r < r_{cr}$, while $s_{av,\infty} = s_3(r) > s_3(r_{cr})$ when $r > r_{cr}$, where $s_1(r)$ and $s_3(r)$, satisfying $s_1(r_{cr}) < s_3(r_{cr})$, are two continuous and strictly increasing functions that depend on $k$, $q$ and $d$.*

**Remark**: Note that random *k*-SAT is a special case of Model GB with $d = 2$ and $q = 1$. It can be verified that for random *k*-SAT the smallest value of $k_N$ in Theorem 1.2 is equal to 4. That is to say, for random *k*-

SAT, the phase transition phenomenon stated in Theorem 1.2 will occur when $k \geq 5$.

The following theorem shows that as $n$ approaches infinity, the expected number of the satisfying assignment pairs with the similarity degree concentrated around $s_{av,\infty}$ is almost equal to the expected number of all the satisfying assignment pairs. In other words, when $n$ is sufficiently large, the typical similarity degree of a satisfying assignment pair is concentrated around $s_{av,\infty}$, implying that $s_{av,\infty}$ is very suitable to characterize how solutions are similar to each other.

**Theorem 1.3**  *Given $r \neq r_{cr}$, for any sufficiently small positive constant $\varepsilon$, we have*

$$\lim_{n \to \infty} \frac{\sum_{s=\frac{[(s_{av,\infty}-\varepsilon)n]+1}{n},\frac{[(s_{av,\infty}-\varepsilon)n]+2}{n},\cdots,\frac{[(s_{av,\infty}+\varepsilon)n]}{n}} E|A_s^{sat}|}{\sum_{s=\frac{0}{n},\frac{1}{n},\cdots,\frac{n}{n}} E|A_s^{sat}|} = 1, \quad (1.5)$$

*where $[a]$ denotes the largest integer smaller than or equal to $a$.*

The layout of this paper is as follows. In Section 2 we will we calculate $E(|A_s^{Sat}|)$ and give an asymptotic estimate of it. Section 3 will investigate how the maximum points of $E(|A_s^{Sat}|)$ vary with $r$ for sufficiently large $n$. The theorems in this paper will be proved in Section 4. In Section 5 we will give an asymptotic estimate of the second moment of the number of solutions for Model GB. Conclusions and future studies will be discussed in Section 6.

## 2. The expected number of satisfying assignment pairs

In this section we will first calculate $E(|A_s^{Sat}|)$ and then give an asymptotic estimate of it for large $n$. Finally, we will use this asymptotic estimate to investigate the properties of $E(|A_s^{Sat}|)$. Recall that $\phi$ is a random CSP instance generated following Model GB and $A_s^{Sat}$ is the set of assignment pairs that are in $A_s$ and satisfy $\phi$. Let $<t_i,t_j>_s$ denote an assignment pair in $A_s$. The probability of $<t_i,t_j>_s$ satisfying $\phi$ is denoted by $P(<t_i,t_j>_s)$. Therefore, $E(|A_s^{Sat}|)$, i.e. the expected value of $|A_s^{Sat}|$ is given by

$$E(|A_s^{Sat}|) = P(<t_i,t_j>_s)|A_s|. \quad (2.1)$$

Now we start to derive $P(<t_i,t_j>_s)$. Since each constraint is generated independently, we only need to consider the probability of $<t_i,t_j>_s$ satisfying a random constraint. Note that the similarity number of $<t_i,t_j>_s$ is equal to $S = ns$, we have the following two cases:

(1) Each variable of a constraint is assigned the same value in $t_i$ as that in $t_j$. In this case, the probability of $<t_i,t_j>$ satisfying the constraint is $\binom{d^k-1}{q}/\binom{d^k}{q}$.

(2) Otherwise, the probability of $<t_i,t_j>$ satisfying a constraint is $\binom{d^k-2}{q}/\binom{d^k}{q}$.

The probability that a random constraint falls into the first case is $\binom{S}{k}/\binom{n}{k}$. Hence the probability into the second case is $1 - \binom{S}{k} / \binom{n}{k}$. Thus we get

$$P(<t_i, t_j>_s) = \left( \frac{\binom{d^k-1}{q}}{\binom{d^k}{q}} \cdot \frac{\binom{S}{k}}{\binom{n}{k}} + \frac{\binom{d^k-2}{q}}{\binom{d^k}{q}} \cdot (1 - \frac{\binom{S}{k}}{\binom{n}{k}}) \right)^{rn}$$

$$= \left( \frac{(d^k-q)[d^k - q - 1 + q \frac{S(S-1)...(S-k+1)}{n(n-1)...(n-k+1)}]}{d^k(d^k-1)} \right)^{rn} \quad (2.2)$$

Note that $S = ns$. As $n$ approaches infinity, we have

$$\frac{S(S-1)(S-2)...(S-k+1)}{n(n-1)(n-2)...(n-k+1)} = s^k + \frac{k(k-1)}{2n}(s^k - s^{k-1}) + O(\frac{1}{n^2}) \quad (2.3)$$

Substituting Eq. (2.3) into Eq. (2.2), we obtain

$$P(<t_i, t_j>_s) = \left( \frac{(d^k-q)[d^k - q - 1 + qs^k + \frac{qk(k-1)}{2n}(s^k - s^{k-1}) + O(\frac{1}{n^2})]}{d^k(d^k-1)} \right)^{rn}. \quad (2.4)$$

Estimating the above equation as $n$ tends to infinity, we deduce

$$P(<t_i, t_j>_s) = \sigma(s) e^{-nrg(s)}(1 + O(\frac{1}{n})) \text{ when } n \to \infty, \quad (2.5)$$

where,

$$\sigma(s) = e^{r\rho(s)}, \quad g(s) = \ln d^k + \ln(d^k - 1) - \ln(d^k - q) - \ln(d^k - q - 1 + qs^k),$$

$$\rho(s) = \frac{qk(k-1)}{2(d^k - q - 1 + qs^k)}(s^k - s^{k-1}). \quad (2.6)$$

Below we will derive and estimate $|A_s|$. By Definitions 1.3 and 1.4 the similarity number of the assignment pairs in $A_s$ is equal to $S = ns$. It is easy to show that the cardinality of $A_s$ is given by

$$|A_s| = d^n \binom{n}{ns}(d-1)^{n-ns}. \quad (2.7)$$

For $s = 0$ or $s = 1$, it is easy to show that

$$|A_s| = d^n(d-1)^n \text{ when } s = 0, \quad (2.8)$$

$$|A_s| = d^n \text{ when } s = 1. \quad (2.9)$$

For every $0 < s < 1$, the asymptotic estimate of $|A_s|$ is

$$|A_s| = \frac{1}{\sqrt{2\pi ns(1-s)}} e^{n[\ln d - s \ln s - (1-s)\ln(1-s) + (1-s)\ln(d-1)]}(1 + O(\frac{1}{n})) \text{ when } n \to \infty. \quad (2.10)$$

Hence we have

$$|A_s| = \tau(s) e^{nh(s)}(1 + O(\frac{1}{n})) \text{ when } n \to \infty, \quad (2.11)$$

where

$$h(s) = \begin{cases} \ln d + (1-s)\ln(d-1) & \text{if } s = 0, 1 \\ \ln d - s\ln s - (1-s)\ln(1-s) + (1-s)\ln(d-1) & \text{if } 0 < s < 1 \end{cases}, \quad \tau(s) = \begin{cases} 1 & \text{if } s = 0, 1 \\ \dfrac{1}{\sqrt{2\pi n s(1-s)}} & \text{if } 0 < s < 1 \end{cases}.$$

(2.12)

By Eqs. (2.1), (2.5) and (2.11) we get

$$E(|A_s^{Sat}|) = \varphi(s) e^{nf(s)}(1 + O(1/n)) \quad \text{when } n \to +\infty,$$  (2.13)

where $\varphi(s)$ and $f(s)$, defined on the interval $0 \leq s \leq 1$, are as follows:

for $0 < s < 1$:

$$\varphi(s) = \sigma(s)\tau(s) = \frac{e^{r\frac{qk(k-1)}{2(d^k - q - 1 + qs^k)}(s^k - s^{k-1})}}{\sqrt{2\pi n s(1-s)}},$$

$$f(s) = h(s) - rg(s)$$
$$= \ln d - s\ln s - (1-s)\ln(1-s) + (1-s)\ln(d-1)$$
$$- r[\ln d^k + \ln(d^k - 1) - \ln(d^k - q) - \ln(d^k - q - 1 + qs^k)];$$  (2.14)

for $s = 0$ or $s = 1$:

$$\varphi(s) = 1,$$

$$f(s) = \ln d + (1-s)\ln(d-1) - r[\ln d^k + \ln(d^k - 1) - \ln(d^k - q) - \ln(d^k - q - 1 + qs^k)].$$  (2.15)

Given $r$, we can obtain the asymptotic estimate of $E(|A_s^{Sat}|)$ for every similarity degree over [0,1] using the above equation. Consequently, it is of interest to find the values of similarity degree maximizing $E(|A_s^{Sat}|)$ for large $n$. First, we have the following definition.

**Definition 2.1** *Given $r$, if $0 \leq s_0 \leq 1$ satisfies the following condition: for every $0 \leq s \leq 1$, there exists $M > 0$ such that $E(|A_s^{Sat}|) \leq E(|A_{s_0}^{Sat}|)$ whenever $n > M$, then $s_0$, denoted by $s_m$ in this paper, is called an asymptotic maximum point of $E(|A_s^{Sat}|)$.*

By Definition 2.1 it is easy to prove that the following propositions hold.

**Proposition 2.1** *Given $r$, if $s_0$ is an asymptotic maximum point of $E(|A_s^{Sat}|)$, then $s_0$ is a maximum point of $f(s)$.*

**Proposition 2.2** *Given $r$, if $s_0$ is the unique maximum point of $f(s)$, then $s_0$ is the unique asymptotic maximum point of $E(|A_s^{Sat}|)$.*

**Proposition 2.3** *Given $r$, if $s_0$ is a maximum point of $f(s)$, then for every $0 \leq s \leq 1$ that is not the maximum point of $f(s)$, there exists $\delta > 0$ and $M > 0$ such that $E(|A_{s_0}^{Sat}|) \geq e^{n\delta} E(|A_s^{Sat}|)$ whenever $n > M$.*

From Proposition 2.1 we know that $s_0$ being a maximum point $f(s)$ is a necessary condition for $s_0$ being an asymptotic maximum point of $E(|A_s^{Sat}|)$. Proposition 2.2 further shows that if $s_0$ is the unique maximum point of $f(s)$, then $s_0$ not only is an asymptotic maximum point of $E(|A_s^{Sat}|)$ but also is the unique asymptotic maximum point. Proposition 2.3 gives us an intuitive understanding of how the values of $s$ maximizing $f(s)$ differs from the other values of $s$ in $E(|A_s^{Sat}|)$.

## 3. The behaviour of $s_m$ as a function of $r$

In this section we will study how the asymptotic maximum points of $E(|A_s^{Sat}|)$ vary with $r$. Note that $f(s)$ is a continuous function over $[0,1]$. The critical points of $f(s)$ satisfy the following equations:

$$f'(s) = h'(s) - rg'(s) = -\ln s + \ln(1-s) - \ln(d-1) + r\frac{kqs^{k-1}}{d^k - q - 1 + qs^k} = 0, \quad (3.1)$$

$$\Leftrightarrow r(s) = \frac{h'(s)}{g'(s)} = \frac{1}{k}(\frac{d^k - q - 1}{qs^{k-1}} + s)[\ln s - \ln(1-s) + \ln(d-1)]. \quad (3.2)$$

**Remark**: Eq. (3.2) is meaningful only for $r(s) \geq 0$. Hence the critical points satisfy $1/d \leq s < 1$.

Eq. (3.2) gives a functional relation between $r$ and the critical points. By examining the behaviour of this function, we can get the relation between $r$ and the maximum points of $f(s)$, and so obtain the behavior of $s_m$ as a function of $r$. To investigate the behaviour of $r(s)$, we first analyze its derivatives.

**Proposition 3.1** *Given $k$, $q$ and $d$, there is only one root of $r''(s) = 0$ over $[1/d, 1)$, denoted by $s_{02}$, and $r''(s) < 0$ when $s < s_{02}$; $r''(s) > 0$ when $s > s_{02}$.*

**Proof.** The second derivative $r''(s)$ of $r(s)$ is as follows:

$$r''(s) = \frac{1}{k}\{\frac{(d^k - q - 1)k(k-1)}{qs^{k+1}}[\ln s - \ln(1-s) + \ln(d-1)] + 2\frac{[qs^k - (d^k - q - 1)(k-1)]}{qs^{k+1}}\frac{1}{(1-s)}$$

$$+ \frac{(d^k - q - 1 + qs^k)}{qs^{k+1}}\frac{(2s-1)}{(1-s)^2}\}. \quad (3.3)$$

Let $F(s) = r''(s)\frac{kqs^{k+1}}{d^k - q - 1}(1-s)^2$. Note that $\frac{kqs^{k+1}}{d^k - q - 1}(1-s)^2 > 0$ over $[1/d, 1)$. Thus $F(s)$ has the same sign with $r''(s)$ over $[1/d, 1)$, i.e. they are both positive, negative or equal to zero at the same point of $s$.

$$F(s) = k(k-1)[\ln s - \ln(1-s) + \ln(d-1)](1-s)^2 + 2ks - 2k + 1 + \frac{qs^k}{d^k - q - 1}. \quad (3.4)$$

Let

$$F_1(s) = k(k-1)[\ln s - \ln(1-s) + \ln(d-1)](1-s)^2 + 2ks - 2k + 1. \quad (3.5)$$

So we have

$$F(s) = F_1(s) + \frac{qs^k}{d^k - q - 1},$$

$$F(\frac{1}{d}) = F_1(\frac{1}{d}) + \frac{q}{d^k(d^k - q - 1)} = 2k(-1 + \frac{1}{d}) + 1 + \frac{q}{d^k(d^k - q - 1)}. \quad (3.6)$$

It is easy to show that $F(1/d) < 0$. The limits of $F(s)$ and $F_1(s)$ as $s \to 1$ are

$$\lim_{s \to 1} F(s) = 1 + \frac{q}{d^k - q - 1} > 0, \quad \lim_{s \to 1} F_1(s) = 1 > 0. \tag{3.7}$$

The first derivatives of $F(s)$ and $F_1(s)$ are

$$F'(s) = F_1'(s) + \frac{qks^{k-1}}{d^k - q - 1}, \tag{3.8}$$

$$F_1'(s) = k(k-1)\frac{1-s}{s} - 2k(k-1)[\ln s - \ln(1-s) + \ln(d-1)](1-s) + 2k. \tag{3.9}$$

Substituting $1/d$ into the above equations, we get

$$F_1'(\frac{1}{d}) = k(k-1)(d-1) + 2k, \quad F'(\frac{1}{d}) = F_1'(\frac{1}{d}) + \frac{qk}{(d^k - q - 1)d^{k-1}}. \tag{3.10}$$

Note that $F(s)$ is a continuous function over $[1/d,1)$. By the intermediate value theorem and Eqs. (3.6), (3.7), there exists at least one root $s_{02}$ such that $F(s_{02}) = 0$. We can further prove that there is at most one root. The proof is divided into the following two cases:

**Case 1.** Assume that there exists no root of equation $F_1'(s) = 0$ over $[1/d,1)$. Since $F_1'(1/d) > 0$, it is easy to see from the assumption that $F_1'(s) > 0$ over $[1/d,1)$. By Eq. (3.8) we get that $F'(s) > 0$, i.e. $F(s)$ is a strictly increasing function over $[1/d,1)$. Thus there is only one root of $F(s) = 0$ over $[1/d,1)$.

**Case 2.** Assume that $s_0$ is a root of $F_1'(s) = 0$ over $[1/d,1)$. So we have

$$k(k-1)\frac{1-s_0}{s_0} - 2k(k-1)[\ln s_0 - \ln(1-s_0) + \ln(d-1)](1-s_0) + 2k = 0. \tag{3.11}$$

Arranging the above equation gives

$$(k-1)[\ln s_0 - \ln(1-s_0) + \ln(d-1)](1-s_0) = (k-1)\frac{1-s_0}{2s_0} + 1. \tag{3.12}$$

Substituting the above equation into Eq. (3.5) yields

$$F_1(s_0) = k(k-1)\frac{(1-s_0)^2}{2s_0} + k(1-s_0) + 2ks_0 - 2k + 1$$

$$= \frac{1}{2s_0}[(k^2 + k)s_0^2 - (2k^2 - 2)s_0 + k(k-1)]. \tag{3.13}$$

Let

$$F_2(s) = (k^2 + k)s^2 - (2k^2 - 2)s + k(k-1). \tag{3.14}$$

Note that $F_2(s)$ is a quadratic function in $s$ and the coefficient of $s^2$ is greater than zero. Its discriminant is

$$\Delta = 4(k^2 - 1)^2 - 4k(k-1)(k^2 + k) = -4(k^2 - 1) < 0. \tag{3.15}$$

Hence we obtain that $F_2(s) > 0$. From $F_1(s_0) = F_2(s_0)/(2s_0)$ we know that $F_1(s_0) > 0$. Recall that $s_0$ is a root of $F_1'(s) = 0$ over $[1/d,1)$, i.e. a critical point of $F_1(s)$ over $[1/d,1)$. Thus $F_1(s)$ is greater than zero at the critical points. Assume that $s_{0a}$ is the smallest critical point. It is easy to show that $F_1'(s) > 0$ over $[1/d, s_{0a})$. Otherwise, if there exists a point satisfying $F_1'(s) \leq 0$ over $[1/d, s_{0a})$, then we can always find a point $s_{00}$ (where $1/d < s_{00} < s_{0a}$) such that $F_1'(s_{00}) = 0$. This is in contradiction with the statement that $s_{01}$ is the smallest critical point. Consequently, we obtain that $F_1'(s) > 0$ over $[1/d, s_{0a})$. By Eq. (3.8) it is easy to prove that $F'(s) > 0$, i.e. $F(s)$ is a strictly increasing function over $[1/d, s_{0a})$. From Eq. (3.6) we know that $F(1/d) < 0$. It follows from Eq. (3.4) that $F(s_{0a}) > 0$. Thus there is only one root of $F(s) = 0$, denoted by $s_{02}$, over $[1/d, s_{0a})$. Note that $F_1(s)$ is greater than zero at the critical points. It is therefore not hard to prove that $F_1(s) > 0$ over $[s_{0a}, 1)$. Hence we deduce that $F(s) > 0$ over $[s_{0a}, 1)$. Combining the above two cases, we obtain that there is only one root $s_{02}$ of $F(s) = 0$ over $[1/d, 1)$, and $F(s) < 0$ when $s < s_{02}$; $F(s) > 0$ when $s > s_{02}$. Recall that $F(s)$ has the same sign with $r''(s)$. Hence Proposition 3.1 is proved.

From Proposition 3.1 we know that $r'(s)$ is a strictly decreasing function over $[1/d, s_{02})$ and a strictly increasing function over $(s_{02}, 1)$. Thus $s_{02}$ is the minimum point of $r'(s)$. In what follows, we will examine the behaviour of $r'(s)$ as $s$ varies.

**Proposition 3.2** *Given $k$ and $q$, there exists $d_N > 0$ such that for any $d > d_N$, there are two and only two roots of $r'(s) = 0$, denoted by $s_{01}$ and $s_{03}$ (where $s_{01} < s_{02} < s_{03}$), over $[1/d, 1)$, and $r'(s) > 0$ on the interval $1/d \leq s < s_{01}$; $r'(s) < 0$ on the interval $s_{01} < s < s_{03}$; $r'(s) > 0$ on the interval $s_{03} < s < 1$.*

**Proof.** The first derivative of $r(s)$ is

$$r'(s) = \frac{1}{k}(1 - \frac{(d^k - q - 1)(k-1)}{qs^k})[\ln s - \ln(1-s) + \ln(d-1)] + \frac{1}{k}(\frac{d^k - q - 1}{qs^{k-1}} + s)(\frac{1}{s} + \frac{1}{1-s}). \quad (3.16)$$

Arranging the above equation, we get

$$r'(s)\frac{kqs^k}{(k-1)(d^k - q - 1) - qs^k} = -[\ln s - \ln(1-s) + \ln(d-1)] + \frac{d^k - q - 1 + qs^k}{(k-1)(d^k - q - 1) - qs^k}\frac{1}{(1-s)}$$

$$\leq -[\ln s - \ln(1-s) + \ln(d-1)] + \frac{d^k - q - 1 + q}{(k-1)(d^k - q - 1) - q}\frac{1}{(1-s)}. \quad (3.17)$$

Let

$$U(s) = \frac{a}{1-s} - [\ln s - \ln(1-s) + \ln(d-1)], \text{ where } a = \frac{d^k - q - 1 + q}{(k-1)(d^k - q - 1) - q}. \quad (3.18)$$

Hence we can write Inequality (3.17) as

$$r'(s)\frac{kqs^k}{(k-1)(d^k - q - 1) - qs^k} \leq U(s). \quad (3.19)$$

It is easy to see that given $k$ and $q$, the parameter $a$ is a strictly decreasing function of $d$. As $d$ tends to infinity, we have

$$\lim_{d \to +\infty} a = \lim_{d \to +\infty} \frac{d^k - q - 1 + q}{(k-1)(d^k - q - 1) - q} = \frac{1}{k-1}, \quad \lim_{d \to +\infty} \frac{1}{d} = 0. \quad (3.20)$$

Note that $k \geq 2$. By the above equations we can easily prove that there exists $d_{N1} > 0$ such that

$$\frac{1}{d} < \frac{1}{1+a} < 1 \text{ when } d > d_{N1}. \quad (3.21)$$

Substituting $1/(1+a)$ into $F(s)$ gives

$$U(\frac{1}{1+a}) = a + 1 + \ln a - \ln(d-1). \quad (3.22)$$

As $d$ tends to infinity, we have

$$\lim_{d\to+\infty}[a+1+\ln a-\ln(d-1)]=-\infty. \tag{3.23}$$

Thus there exists $d_{N2}>0$ such that

$$U(\frac{1}{1+a})<0 \text{ when } d>d_{N2}. \tag{3.24}$$

By Eqs. (3.19) and (3.24) we get

$$r'(\frac{1}{1+a})\frac{kq}{(k-1)(d^k-q-1)(1+a)^k-q}\leq U(\frac{1}{1+a})<0. \tag{3.25}$$

Recall that $d\geq 2$ and $q<d^{k-1}$ in Model GB. We can easily deduce

$$\frac{kq}{(k-1)(d^k-q-1)(1+a)^k-q}>0. \tag{3.26}$$

Let $d_N=\max\{d_{N1},d_{N2}\}$. From Proposition 3.1 we know that $s_{02}$ is the minimum point of $r'(s)$. So we find

$$r'(s_{02})\leq r'(\frac{1}{1+a})<0 \text{ when } d>d_N. \tag{3.27}$$

By Eq. (3.16) we obtain

$$r'(\frac{1}{d})=\frac{1}{k}[\frac{d^{k-1}(d^k-q-1)}{q}+\frac{1}{d}]\frac{d^2}{d-1}>0,\ \lim_{s\to 1}r'(s)=+\infty. \tag{3.28}$$

From Proposition 3.1 we know that $r'(s)$ is a continuous and strictly decreasing function over $[1/d,s_{02})$. Thus there exists $s_{01}$ such that $r'(s_{01})=0$, and $r'(s)>0$ when $1/d\leq s<s_{01}$; $r'(s)<0$ when $s_{01}<s<s_{02}$. Similarly, we can easily prove that there exists $s_{03}$ such that $r'(s_{03})=0$, and $r'(s)<0$ when $s_{02}<s<s_{03}$; $r'(s)>0$ when $s_{03}<s<1$. Hence Proposition 3.2 is proved.

Similar to Proposition 3.2, we can easily obtain the following proposition.

**Proposition 3.3** *Given $d$ and $q$, there exists $k_N>0$ such that for any $k>k_N$, there are two and only two roots of $r'(s)=0$, denoted by $s_{01}$ and $s_{03}$ (where $s_{01}<s_{02}<s_{03}$), over $[1/d,1)$, and $r'(s)>0$ on the interval $1/d\leq s<s_{01}$; $r'(s)<0$ on the interval $s_{01}<s<s_{03}$; $r'(s)>0$ on the interval $s_{03}<s<1$.*

**Proof.** By Eq. (3.18) it is easy to show that given $d$ and $q$, the parameter $a$ is a strictly decreasing function of $k$. As $k$ approaches infinity, we have

$$\lim_{k\to+\infty}a=\lim_{k\to+\infty}\frac{d^k-q-1+q}{(k-1)(d^k-q-1)-q}=0. \tag{3.29}$$

It is straightforward from the above equation that there exists $k_{N1}>0$ such that

$$\frac{1}{d}<\frac{1}{1+a}<1 \text{ when } k>k_{N1}. \tag{3.30}$$

Substituting $1/(1+a)$ into $U(s)$ yields

$$U(\frac{1}{1+a})=a+1+\ln a-\ln(d-1). \tag{3.31}$$

As $k$ approaches infinity, we have

$$\lim_{k\to+\infty}[a+1+\ln a-\ln(d-1)]=-\infty. \tag{3.32}$$

Thus there exists $k_{N2} > 0$ such that

$$U(\frac{1}{1+a}) < 0 \text{ when } k > k_{N2}. \tag{3.33}$$

Let $k_N = \max\{k_{N1}, k_{N2}\}$. We have

$$r'(s_{02}) \le r'(\frac{1}{1+a}) < 0 \text{ when } k > k_N. \tag{3.34}$$

The remainder of the proof is along the same lines as those in the proof Proposition 3.2. Recall that random $k$-SAT is a special case of Model GB with $d = 2$ and $q = 1$. It can be verified that for random $k$-SAT, Proposition 3.2 holds when $k \ge 5$ [22].

The following picture gives us an intuitive understanding of how the function $r(s)$ varies with $s$ for given values of $k$, $q$ and $d$ satisfying the conditions in Propositions 3.2 or 3.3.

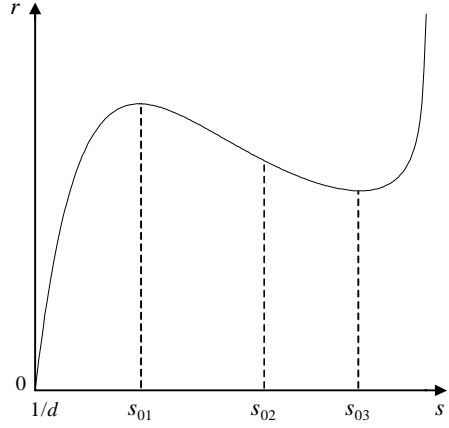

Fig. 1  The curve of $r(s)$ as a function of $s$.

From the above picture we know that $r(s)$ is a strictly increasing function on the intervals $1/d \le s \le s_{01}$ and $s_{03} \le s < 1$, and is a strictly decreasing function on the interval $s_{01} \le s \le s_{03}$. So we can define the inverse functions of $r(s)$ in every interval as:

$$s_1(r) = r^{-1}(s): \ [0, r(s_{01})] \to [1/d, s_{01}];$$

$$s_2(r) = r^{-1}(s): \ [r(s_{03}), r(s_{01})] \to [s_{01}, s_{03}];$$

$$s_3(r) = r^{-1}(s): \ [r(s_{03}), +\infty) \to [s_{03}, 1).$$

We know from Eq. (3.2) that $r(s)$ depends on $k$, $q$ and $d$. Consequently, if the values of $k$, $q$ and $d$ satisfying the conditions in Propositions 3.2 or 3.3 are given, then the functions $s_1(r)$, $s_2(r)$ and $s_3(r)$ can be exactly determined.

**Proposition 3.4**  *Given $k$, $q$, there exists $d_N > 0$ such that for every $d > d_N$, there exists a threshold point $r_{cr}$ (the value of $r_{cr}$ depends on $k$, $q$ and $d$) such that the unique asymptotic maximum point is $s_m = s_1(r) < s_1(r_{cr})$ when $0 < r < r_{cr}$, while the unique asymptotic maximum point is $s_m = s_3(r) > s_3(r_{cr})$ when $r > r_{cr}$, where $s_1(r)$ and $s_3(r)$, satisfying $s_1(r_{cr}) < s_3(r_{cr})$, are two continuous and strictly increasing functions that depend on $k$, $q$ and $d$.*

**Proof.** Given $r_0 < r(s_{03})$, it is easy to see from Fig. 1 that there exists only one critical point $s_1(r_0)$ of $f(s)$. By Eqs. (3.1), (3.2) and $r_0 = h'(s_1(r_0))/g'(s_1(r_0))$, we get

$$r'(s_1(r_0)) = \frac{h''(s_1(r_0))g'(s_1(r_0)) - h'(s_1(r_0))g''(s_1(r_0))}{(g'(s_1(r_0)))^2}, \quad g'(s_1(r_0)) = -\frac{kq \cdot (s_1(r_0))^{k-1}}{d^k - q - 1 + q \cdot (s_1(r_0))^k}, \quad (3.35)$$

$$f''(s_1(r_0)) = h''(s_1(r_0)) - r_0 g''(s_1(r_0)) = \frac{h''(s_1(r_0))g'(s_1(r_0)) - h'(s_1(r_0))g''(s_1(r_0))}{g'(s_1(r_0))}. \quad (3.36)$$

By Proposition 3.2 we get that $r'(s_1(r_0)) > 0$. It follows from Eq. (3.35) that $g'(s_1(r_0)) < 0$. Hence we have

$$f''(s_1(r_0)) = r'(s_1(r_0))g'(s_1(r_0)) < 0. \quad (3.37)$$

Therefore, given $r_0 < r(s_{03})$, $s_1(r_0)$ is the unique maximum point of $f(s)$. By Proposition 3.1 $s_1(r_0)$ is also the unique asymptotic maximum point of $E(|A_s^{Sat}|)$. Similarly, given $r_0 > r(s_{01})$, $s_3(r_0)$ is the unique asymptotic maximum point $E(|A_s^{Sat}|)$. Notice that $s_1(r)$ and $s_3(r)$ are continuous and strictly increasing functions. Therefore, if $r < r(s_{03})$ or $r > r(s_{01})$, the asymptotic maximum point of $E(|A_s^{Sat}|)$ will increase continuously as $r$ grows. However, it will be proved below that the asymptotic maximum point $s_m$ is discontinuous at a threshold point on the interval $r(s_{03}) \leq r \leq r(s_{01})$ as $r$ varies. Given $r(s_{03}) < r_0 < r(s_{01})$, there are three critical points of $f(s)$ which are $s_1(r_0)$, $s_2(r_0)$ and $s_3(r_0)$ respectively. Similarly, by Eq. (3.36) it can be easily deduced that $f''(s_1(r_0)) > 0$, $f''(s_3(r_0)) > 0$ and $f''(s_2(r_0)) < 0$. Hence $s_1(r_0)$ and $s_3(r_0)$ are local maximum points while $s_2(r_0)$ is a local minimum point. The maximum point can be easily obtained by deciding which local maximum value is greater. We first define the following function:

$$F(r) = f(s_1(r)) - f(s_3(r)), \quad r(s_{03}) \leq r \leq r(s_{01}). \quad (3.38)$$

By Fig. 1 and Eq. (3.36), if $r = r(s_{03})$, then $s_1(r(s_{03}))$ is the sole maximum point of $f(s)$. Hence we have

$$F(r(s_{03})) = f(s_1(r(s_{03}))) - f(s_3(r(s_{03}))) > 0. \quad (3.39)$$

Similarly, if $r = r(s_{01})$, then $s_3(r(s_{01}))$ is the only one maximum point of $f(s)$. Thus we deduce

$$F(r(s_{01})) = f(s_1(r(s_{01}))) - f(s_3(r(s_{01}))) < 0. \quad (3.40)$$

The first derivative of $F(r)$ is as follows

$$F'(r) = h'(s_1(r))s_1'(r) - g(s_1(r)) - rg'(s_1(r))s_1'(r) - h'(s_3(r))s_3'(r) + g(s_3(r)) + rg'(s_3(r))s_3'(r). \quad (3.41)$$

By use of the condition that $s_1(r)$ and $s_3(r)$ are the critical points of $f(s)$, we have

$$h'(s_1(r)) - rg'(s_1(r)) = 0, \quad h'(s_3(r)) - rg'(s_3(r)) = 0. \quad (3.42)$$

Substituting Eq. (3.42) into Eq. (3.41), we obtain

$$F'(r) = \ln(d^k - q - 1 + q \cdot (s_1(r))^k) - \ln(d^k - q - 1 + q \cdot (s_3(r))^k). \quad (3.43)$$

It is obvious that $s_1(r) \leq s_{01} < s_{03} \leq s_3(r)$. Thus

$$F'(r) < 0. \quad (3.44)$$

By the intermediate value theorem and Eqs. (3.39), (3.40) and (3.44), there exists only one root $r_{cr}$ of $F(r) = 0$, and the following facts hold:

If $r(s_{03}) \leq r < r_{cr}$, then $f(s_1(r)) > f(s_3(r))$. Hence $s_m = s_1(r) < s_1(r_{cr})$.

If $r_{cr} < r \leq r(s_{01})$, then $f(s_1(r)) < f(s_3(r))$. Hence $s_m = s_3(r) > s_3(r_{cr})$.

It is easy to see that $s_1(r_{cr}) < s_3(r_{cr})$. Hence Proposition 3.4 is proved.

Along the same lines as those in the proof Proposition 3.4, we can easily obtain the following proposition.

**Proposition 3.5** *Given $d$, $q$, there exists $k_N > 0$ such that for every $k > k_N$, there exists a threshold point $r_{cr}$ (the value of $r_{cr}$ depends on $k$, $q$ and $d$) such that the unique asymptotic maximum point is $s_m = s_1(r) < s_1(r_{cr})$ when $0 < r < r_{cr}$, while the unique asymptotic maximum point is $s_m = s_3(r) > s_3(r_{cr})$ when $r > r_{cr}$, where $s_1(r)$ and $s_3(r)$, satisfying $s_1(r_{cr}) < s_3(r_{cr})$, are two continuous and strictly increasing functions that depend on $k$, $q$ and $d$.*

## 4. Proofs of theorems

In this section, we will prove Theorems 1.1 and 1.2 on the phase transition behaviour of the average similarity degree and Theorem 1.3 on the concentration of similarity degree around $s_{av,\infty}$. From Definition 1.4 we know that calculating the average similarity degree involves summing the terms $E(|A_s^{Sat}|)$ over $s = 0/n, 1/n, \cdots, n/n$. Recall that the asymptotic estimate of $E(|A_s^{Sat}|)$ can be written as $\varphi(s)e^{nf(s)}$. The following proposition gives a method of how to sum the terms with the form of $\varphi(s)e^{nf(s)}$ over a fixed interval.

**Proposition 4.1** *Assume that $\varphi(x)$ and $f(x)$ are two bounded functions defined over $[\alpha, \beta]$ where $\alpha < \beta$ are two constants, and satisfy the following conditions:*

*(1) There is only one maximum point $\zeta$ of $f(x)$ over $[\alpha, \beta]$ where $\zeta \in (\alpha, \beta)$, and $f''(\zeta) < 0$.*

*(2) $f''(x)$ is continuous and $f'''(x)$ exists in some neighborhood of $\zeta$.*

*(3) $\varphi(\zeta) \neq 0$, and $\varphi(x)$ is continuous at $x = \zeta$.*

*Then we have*

$$\sum_{i=[n\alpha]+1}^{[n\beta]} \varphi(\frac{i}{n}) e^{nf(\frac{i}{n})} = \varphi(\zeta) \sqrt{\frac{2n\pi}{-f''(\zeta)}} e^{nf(\zeta)} (1+o(1)) \quad \text{when} \quad n \to \infty.$$

**Proof.** From condition (1) we know that $f''(\zeta) < 0$. Let $\delta$ be a sufficiently small positive constant such that

$$f''(x) \leq -s < 0, \quad x \in [\zeta - \delta, \zeta + \delta]. \tag{4.1}$$

Using a technique similar to Laplace's method for integrals[15], we have

$$\sum_{i=[n\alpha]+1}^{[n\beta]} \varphi(\frac{i}{n}) e^{n(f(\frac{i}{n})-f(\zeta))} = I_1 + I_2 + I_3 + I_4 + I_5, \tag{4.2}$$

where $I_1 = \sum_{i=[n\alpha]+1}^{[n(\zeta-\delta)]} Q(\frac{i}{n})$, $I_2 = \sum_{i=[n(\zeta-\delta)]+1}^{[(n\zeta-n^{\frac{3}{5}})]} Q(\frac{i}{n})$, $I_3 = \sum_{i=[(n\zeta-n^{\frac{3}{5}})]+1}^{[(n\zeta+n^{\frac{3}{5}})]} Q(\frac{i}{n})$, $I_4 = \sum_{i=[(n\zeta+n^{\frac{3}{5}})]+1}^{[n(\zeta+\delta)]} Q(\frac{i}{n})$, $I_5 = \sum_{i=[n(\zeta+\delta)]+1}^{[n\beta]} Q(\frac{i}{n})$,

$$Q(\frac{i}{n}) = \varphi(\frac{i}{n})e^{n(f(\frac{i}{n})-f(\zeta))}.$$

Let $t_1 = \sup_{x\in[\alpha,\zeta-\delta]} f(x) - f(\zeta)$. By condition (1) we know that $t_1 < 0$. So

$$I_1 = \sum_{i=[n\alpha]+1}^{[n(\zeta-\delta)]} \varphi(\frac{i}{n})e^{n(f(\frac{i}{n})-f(\zeta))} = O(\sum_{i=[n\alpha]+1}^{[n(\zeta-\delta)]} \varphi(\frac{i}{n})e^{nt_1}) = O(ne^{nt_1}). \quad (4.3)$$

Similarly, let $t_5 = \sup_{x\in[\zeta+\delta,\beta]} f(x) - f(\zeta) < 0$. Then

$$I_5 = O(ne^{nt_5}). \quad (4.4)$$

An application of Taylor's Theorem yields

$$f(x) - f(\zeta) = \frac{1}{2}f''(\eta)(x-\zeta)^2, \quad \eta = \zeta + \theta(x-\zeta), \quad 0 < \theta < 1. \quad (4.5)$$

Hence

$$I_2 = \sum_{i=[n(\zeta-\delta)]+1}^{[(n\zeta-n^{\frac{3}{5}})]} \varphi(\frac{i}{n})e^{n(f(\frac{i}{n})-f(\zeta))} = O(\sum_{i=[n(\zeta-\delta)]+1}^{[(n\zeta-n^{\frac{3}{5}})]} \varphi(\frac{i}{n})e^{-n\frac{s}{2}(\frac{i}{n}-\zeta)^2}) = O(\sum_{i=[n(\zeta-\delta)]+1}^{[(n\zeta-n^{\frac{3}{5}})]} \varphi(\frac{i}{n})e^{-\frac{s}{2}n^{\frac{1}{5}}}) = O(ne^{-\frac{s}{2}n^{\frac{1}{5}}}). \quad (4.6)$$

Similarly, we get

$$I_4 = O(ne^{-\frac{s}{2}n^{\frac{1}{5}}}). \quad (4.7)$$

We now start to estimate $I_3$.

Let $i = (n\zeta + l)$ and $l = o(n)$. Expanding $f(x)$ in the Taylor Series about $\zeta$, we obtain

$$f(\zeta + \frac{l}{n}) = f(\zeta) + f''(\zeta)\frac{l^2}{2n^2} + O(\frac{l^3}{n^3}), \quad \varphi(\zeta + \frac{l}{n}) = \varphi(\zeta)(1+o(1)),$$

$$I_3 = \sum_{l=[-n^{\frac{3}{5}}]+1}^{[n^{\frac{3}{5}}]} \varphi(\zeta + \frac{l}{n})e^{n(f(\zeta+\frac{l}{n})-f(\zeta))} = \sum_{l=[-n^{\frac{3}{5}}]+1}^{[n^{\frac{3}{5}}]} \varphi(\zeta + \frac{l}{n})e^{f''(\zeta)\frac{l^2}{2n}+O(\frac{l^3}{n^2})} = (1+o(1))\varphi(\zeta) \sum_{l=[-n^{\frac{3}{5}}]+1}^{[n^{\frac{3}{5}}]} e^{f''(\zeta)\frac{l^2}{2n}}. \quad (4.8)$$

If $-n \leq l \leq -n^{\frac{3}{5}}$ or $n^{\frac{3}{5}} \leq l \leq n$, $e^{f''(\zeta)\frac{l^2}{2n}}$ is exponentially smaller than that of $l = 0$. So we can write Eq. (4.8) as

$$I_3 = (1+o(1))\varphi(\zeta)\sum_{l=-n}^{n} e^{f''(\zeta)\frac{l^2}{2n}}. \quad (4.9)$$

Let $c = -f''(\zeta)$ and $H(x) = e^{-\frac{cx^2}{2n}}$. Applying Euler's summation formula (see [12], p.160 and [19]) to $\sum_{l=-n}^{n} H(l)$, we have

$$\sum_{l=-n}^{n} H(l) = (1+o(1))\int_{-\infty}^{\infty} H(x)dx = (1+o(1))\sqrt{\frac{2n\pi}{c}}. \quad (4.10)$$

Combining the above results gives

$$\sum_{i=[n\alpha]+1}^{[n\beta]} \varphi(\frac{i}{n})e^{n(f(\frac{i}{n})-f(\zeta))} = \varphi(\zeta)\sqrt{\frac{2n\pi}{-f''(\zeta)}}(1+o(1))+O(ne^{nt_1})+O(ne^{nt_5})+O(ne^{-\frac{s}{2}n^{\frac{1}{5}}})$$

$$= \varphi(\zeta)\sqrt{\frac{2n\pi}{-f''(\zeta)}}(1+o(1)) \text{ when } n \to \infty. \qquad (4.11)$$

Multiplying both sides of the above equation by $e^{nf(\zeta)}$, we obtain

$$\sum_{i=[n\alpha]+1}^{[n\beta]} \varphi(\frac{i}{n})e^{nf(\frac{i}{n})} = \varphi(\zeta)\sqrt{\frac{2n\pi}{-f''(\zeta)}} e^{nf(\zeta)}(1+o(1)) \text{ when } n \to \infty.$$

Hence Proposition 4.1 is proved.

It is easy to see from Eq. (2.14) that $\varphi(s)$ tends to infinity as $s$ approaches $s=0$ or $s=1$. So Proposition 4.1 can not be directly used to sum the terms $E(|A_s^{Sat}|)$ over $s=0/n, 1/n, \cdots, n/n$. The following proposition shows that $E(|A_s^{Sat}|)$ increases with $s$ when $s$ is very close to 0, but decreases with $s$ when $s$ is very close to 1. In what follows, this proposition will be used to estimate from above the sum of the terms close to $s=0$ or close to $s=1$.

**Proposition 4.2** *Given $r$, for sufficiently large $n$ there exists two positive constants $0 < \delta_1 < \delta_2 < 1$ such that $E(|A_s^{Sat}|)$ is a strictly increasing function of $s$ over $0 \le s < \delta_1$, and is a strictly decreasing function of $s$ over $\delta_2 < s \le 1$.*

**Proof.** It follows from Eqs. (2.1), (2.2) and (2.7) that

$$\frac{E(|A_{s+\frac{1}{n}}^{Sat}|)}{E(|A_s^{Sat}|)} = \frac{1}{d-1} \frac{n-ns}{ns+1} \left( \frac{d^k-q-1+q\left(\frac{ns}{n}+\frac{1}{n}\right)\left(\frac{ns-1}{n-1}+\frac{1}{n-1}\right)\cdots\left(\frac{ns-k+1}{n-k+1}+\frac{1}{n-k+1}\right)}{d^k-q-1+q\frac{ns}{n}\frac{(ns-1)}{(n-1)}\cdots\frac{(ns-k+1)}{(n-k+1)}} \right)^{r_n}. \qquad (4.12)$$

By the above equation it is easy to show that there exists a positive constant $0 < \delta_1 < 1$ such that $E(|A_{s+\frac{1}{n}}^{Sat}|) > E(|A_s^{Sat}|)$ whenever $s \le \delta_1$. So $E(|A_s^{Sat}|)$ is a strictly increasing function of $s$ over $0 \le s < \delta_1$.

We will examine below how $E(|A_s^{Sat}|)$ varies with $s$ when $s$ is very close to 1. Notice that $k$, denoting the number of variables in a constraint, is a constant. By use of the inequality $\frac{S-l}{n-l} \le \frac{S-l+1}{n-l+1}$ where $1 \le l \le k$, for sufficiently large $n$, we have

$$\frac{ns}{n}\frac{(ns-1)}{(n-1)}\cdots\frac{(ns-k+1)}{(n-k+1)} \ge \left(\frac{ns-k+1}{n-k+1}\right)^k \ge \left(\frac{ns-k+1}{n}\right)^k > \left(s-\frac{k}{n}\right)^k,$$

$$\left(\frac{ns}{n}+\frac{1}{n}\right)\left(\frac{ns-1}{n-1}+\frac{1}{n-1}\right)\cdots\left(\frac{ns-k+1}{n-k+1}+\frac{1}{n-k+1}\right) < \left(s+\frac{1}{n-k+1}\right)^k < \left(s+\frac{2}{n}\right)^k. \qquad (4.13)$$

Combining Eq. (4.12) and Inequality (4.13), we get

$$\frac{E(|A^{Sat}_{s+\frac{1}{n}}|)}{E(|A^{Sat}_s|)} < \frac{1}{d-1}\frac{n-ns}{ns+1}\left(1+\frac{q\left(s+\frac{2}{n}\right)^k - q\left(s-\frac{k}{n}\right)^k}{d^k-q-1}\right)^{rn}. \tag{4.14}$$

Note that $0 \le s \le 1$. It is not hard to prove that the following two inequalities hold

$$\left(s+\frac{2}{n}\right)^k - \left(s-\frac{k}{n}\right)^k \le \left(1+\frac{2}{n}\right)^k - \left(1-\frac{k}{n}\right)^k, \quad \left(1-\frac{k}{n}\right)^k \ge 1-\frac{k^2}{n}, \tag{4.15}$$

and when $n$ is sufficiently large, the following inequality also holds

$$\left(1+\frac{2}{n}\right)^k < 1+\frac{4k}{n}. \tag{4.16}$$

By Inequalities (4.15) and (4.16), we have

$$\left(s+\frac{2}{n}\right)^k - \left(s-\frac{k}{n}\right)^k < \frac{(k^2+4k)}{n}. \tag{4.17}$$

Hence

$$\left(1+\frac{q\left(s+\frac{2}{n}\right)^k - q\left(s-\frac{k}{n}\right)^k}{d^k-q-1}\right)^{rn} < \left(1+\frac{q(k^2+4k)}{d^k-q-1}\frac{1}{n}\right)^{rn}. \tag{4.18}$$

Let $c_1 = \frac{q(k^2+4k)}{d^k-q-1}$. By use of the inequality $\left(1+\frac{1}{x}\right)^x \le e$ where $x>0$, we obtain

$$\left(1+\frac{c_1}{n}\right)^{rn} = \left(1+\frac{c_1}{n}\right)^{\frac{n}{c_1}c_1 r} \le e^{c_1 r}. \tag{4.19}$$

Substituting Inequalities (4.18) and (4.19) into Inequality (4.14) gives

$$\frac{E(|A^{Sat}_{s+\frac{1}{n}}|)}{E(|A^{Sat}_s|)} < \frac{1}{d-1}\frac{n-ns}{ns+1}e^{c_1 r}. \tag{4.20}$$

Let $0 < \delta_2 < 1$ be a positive constant satisfying $\delta_2 > \max(\frac{e^{c_1 r}}{e^{c_1 r}+d-1}, \delta_1)$. From the above inequality we can easily prove that $E(|A^{Sat}_{s+\frac{1}{n}}|) < E(|A^{Sat}_s|)$ whenever $\delta_2 < s \le 1$, i.e. $E(|A^{Sat}_s|)$ is a strictly decreasing function of $s$ over $\delta_2 < s \le 1$. Hence Proposition 4.2 is proved.

The following proposition will establish a connection between the limit of average similarity degree and the unique asymptotic maximum point of $E(|A^{Sat}_s|)$, and thus help us to prove the theorems in this paper.

**Proposition 4.3** *Given $r \ne r_{cr}$, the limit of average similarity degree is equal to the unique asymptotic maximum point of $E(|A^{Sat}_s|)$, i.e. $s_{av,\infty} = s_m$.*

**Proof.** By Eq. (2.13) we have

$$\sum_{s=\frac{0}{n},\frac{1}{n},\cdots,\frac{n}{n}} E\left|A_s^{Sat}\right| \cong \sum_{s=\frac{0}{n},\frac{1}{n},\cdots,\frac{n}{n}} \varphi(s)e^{nf(s)}. \tag{4.21}$$

Let $\alpha_1$ and $\alpha_2$ be two rationals satisfying $0 < \alpha_1 < \delta_1$ and $\delta_2 < \alpha_2 < 1$. From Proposition 4.2 we know that $E(|A_s^{Sat}|)$ is a strictly increasing function of $s$ over $0 \le s < \delta_1$, and is a strictly decreasing function of $s$ over $\delta_2 < s \le 1$. Hence,

$$\sum_{s=\frac{0}{n},\frac{1}{n},\cdots,\frac{\alpha_1 n}{n}} E\left|A_s^{Sat}\right| \le (\alpha_1 n + 1)E\left|A_{s=\alpha_1}^{Sat}\right|. \tag{4.22}$$

It follows from Eq. (2.13) that

$$E\left|A_{s=\alpha_1}^{Sat}\right| \cong \varphi(\alpha_1)e^{nf(\alpha_1)}. \tag{4.23}$$

Consequently,

$$\sum_{s=\frac{0}{n},\frac{1}{n},\cdots,\frac{\alpha_1 n}{n}} E\left|A_s^{Sat}\right| \cong O(ne^{nf(\alpha_1)}). \tag{4.24}$$

Similarly, we get

$$\sum_{s=\frac{\alpha_2 n+1}{n},\frac{\alpha_2 n+2}{n},\cdots,\frac{n}{n}} E\left|A_s^{Sat}\right| \cong O(ne^{nf(\alpha_2)}). \tag{4.25}$$

It is easy to see from Eq. (2.14) that $\varphi(s)$ and $f(s)$ are two continuous functions over $[\alpha_1, \alpha_2]$. From the proof of Proposition 3.4 we know that given $r \ne r_{cr}$, $s_m$ is the unique maximum point of $f(s)$. By Proposition 4.1 we have

$$\sum_{s=\frac{\alpha_1 n+1}{n},\frac{\alpha_1 n+2}{n},\cdots,\frac{\alpha_2 n}{n}} E\left|A_s^{Sat}\right| \cong \varphi(s_m)\sqrt{\frac{2n\pi}{-f''(s_m)}}e^{nf(s_m)}. \tag{4.26}$$

Combining the above cases gives

$$\sum_{s=\frac{0}{n},\frac{1}{n},\cdots,\frac{n}{n}} E\left|A_s^{Sat}\right| \cong \varphi(s_m)\sqrt{\frac{2n\pi}{-f''(s_m)}}e^{nf(s_m)} + O(ne^{nf(\alpha_1)}) + O(ne^{nf(\alpha_2)}) \cong \varphi(s_m)\sqrt{\frac{2n\pi}{-f''(s_m)}}e^{nf(s_m)}. \tag{4.27}$$

Similarly, we can easily deduce

$$\sum_{s=\frac{0}{n},\frac{1}{n},\cdots,\frac{n}{n}} sE\left|A_s^{Sat}\right| \cong s_m\varphi(s_m)\sqrt{\frac{2n\pi}{-f''(s_m)}}e^{nf(s_m)}. \tag{4.28}$$

By Eqs. (1.4), (4.26) and (4.28) we have

$$s_{av,\infty} = \lim_{n \to \infty} \frac{\sum_{s=\frac{0}{n},\frac{1}{n},\cdots,\frac{n}{n}} sE\left|A_s^{Sat}\right|}{\sum_{s=\frac{0}{n},\frac{1}{n},\cdots,\frac{n}{n}} E\left|A_s^{Sat}\right|} = s_m. \tag{4.29}$$

Hence Proposition 4.3 is proved.

We can now easily prove Theorems 1.1, 1.2 and 1.3.

**Proof of Theorem 1.1.** By Propositions 3.4 and 4.3 the proof is straightforward.

**Proof of Theorem 1.2.** By Propositions 3.5 and 4.3 the proof is straightforward.

**Proof of Theorem 1.3.** Given $r \neq r_{cr}$, for any small positive constant $\varepsilon$, it is easy to see from Proposition 3.4 that $s_m$ is the unique maximum point of $f(s)$ over $[s_m - \varepsilon, s_m + \varepsilon]$. By Proposition 4.1 we obtain

$$\sum_{s = \frac{[(s_m-\varepsilon)n]+1}{n}, \frac{[(s_m-\varepsilon)n]+2}{n}, \cdots, \frac{[(s_m+\varepsilon)n]}{n}} E\left|A_s^{Sat}\right| \cong \varphi(s_m) \sqrt{\frac{2n\pi}{-f''(s_m)}} e^{nf(s_m)}. \tag{4.30}$$

It follows from Eq. (4.26) that

$$\sum_{s=\frac{0}{n},\frac{1}{n},\cdots,\frac{n}{n}} E\left|A_s^{Sat}\right| \cong \varphi(s_m) \sqrt{\frac{2n\pi}{-f''(s_m)}} e^{nf(s_m)}. \tag{4.31}$$

By Eqs. (4.30) and (4.31) we have

$$\lim_{n \to \infty} \frac{\sum_{s=\frac{[(s_m-\varepsilon)n]+1}{n},\frac{[(s_m-\varepsilon)n]+2}{n},\cdots,\frac{[(s_m+\varepsilon)n]}{n}} E\left|A_s^{Sat}\right|}{\sum_{s=\frac{0}{n},\frac{1}{n},\cdots,\frac{n}{n}} E\left|A_s^{Sat}\right|} = 1. \tag{4.32}$$

From Proposition 4.3 we know that given $r \neq r_{cr}$, the limit of average similarity degree $s_{av,\infty}$ is equal to the unique asymptotic maximum point $s_m$. Consequently,

$$\lim_{n \to \infty} \frac{\sum_{s=\frac{[(s_{av,\infty}-\varepsilon)n]+1}{n},\frac{[(s_{av,\infty}-\varepsilon)n]+2}{n},\cdots,\frac{[(s_{av,\infty}+\varepsilon)n]}{n}} E\left|A_s^{Sat}\right|}{\sum_{s=\frac{0}{n},\frac{1}{n},\cdots,\frac{n}{n}} E\left|A_s^{Sat}\right|} = 1. \tag{4.33}$$

Hence Theorem 1.3 is proved.

## 5. The second moment of the number of solutions

In this section, we will derive the second moment of the number of solutions, denoted by $E(N^2)$ in this paper, for random instances generated following Model GB and give an asymptotic estimate of it. It will be shown that the threshold point $r = r_{cr}$, where the limit of average similarity degree $s_{av,\infty}$ shifts from a smaller value to a larger value abruptly, is also a singular point with respect to $r$ in the asymptotic estimate of $E(N^2)$.

**Proposition 5.1** *We have*
*for $0 < r < r_{cr}$:*

$$E(N^2) \cong \varphi(s_1(r)) \sqrt{\frac{2n\pi}{-f''(s_1(r))}} e^{nf(s_1(r))};$$

*for $r > r_{cr}$:*

$$E(N^2) \cong \varphi(s_3(r)) \sqrt{\frac{2n\pi}{-f''(s_3(r))}} e^{nf(s_3(r))};$$

*for $r = r_{cr}$:*

$$E(N^2) \cong \varphi(s_1(r_{cr}))\sqrt{\frac{2n\pi}{-f''(s_1(r_{cr}))}}e^{nf(s_1(r_{cr}))} + \varphi(s_3(r_{cr}))\sqrt{\frac{2n\pi}{-f''(s_3(r_{cr}))}}e^{nf(s_3(r_{cr}))}.$$

**Proof.** The second moment of the number of solutions for random instances generated following Model GB is given by

$$E(N^2) = \sum_{s=\frac{0}{n},\frac{1}{n},\cdots,\frac{n}{n}} E\left|A_s^{Sat}\right|. \tag{5.1}$$

From Propositions 3.4 and 3.5 we know that for every $0 < r < r_{cr}$ the unique maximum point of $f(s)$ is $s_1(r)$, and for every $r > r_{cr}$ the unique maximum point of $f(s)$ is $s_3(r)$. Hence it follows from Eqs. (4.27) and (5.1) that

for $0 < r < r_{cr}$:

$$E(N^2) \cong \varphi(s_1(r))\sqrt{\frac{2n\pi}{-f''(s_1(r))}}e^{nf(s_1(r))}; \tag{5.2}$$

for $r > r_{cr}$:

$$E(N^2) \cong \varphi(s_3(r))\sqrt{\frac{2n\pi}{-f''(s_3(r))}}e^{nf(s_3(r))}. \tag{5.3}$$

For $r = r_{cr}$, there are two and only two maximum points of $f(s)$ that are $s_1(r_{cr})$ and $s_3(r_{cr})$ respectively. Let $t_0 = \frac{s_1(r_{cr}) + s_3(r_{cr})}{2}$. It is easy to show that $s_1(r_{cr})$ is the unique maximum point over the interval $[0, t_0]$, and $s_3(r_{cr})$ is the unique maximum point over the interval $[t_0, 1]$. We can write Eq. (5.1) as

$$E(N^2) = \sum_{s=\frac{0}{n},\frac{1}{n},\cdots,\frac{[t_0 n]}{n}} E\left|A_s^{Sat}\right| + \sum_{s=\frac{[t_0 n]+1}{n},\frac{[t_0 n]+2}{n},\cdots,\frac{n}{n}} E\left|A_s^{Sat}\right|. \tag{5.4}$$

So we have

$$E(N^2) \cong \varphi(s_1(r_{cr}))\sqrt{\frac{2n\pi}{-f''(s_1(r_{cr}))}}e^{nf(s_1(r_{cr}))} + \varphi(s_3(r_{cr}))\sqrt{\frac{2n\pi}{-f''(s_3(r_{cr}))}}e^{nf(s_3(r_{cr}))}. \tag{5.5}$$

Hence Proposition 5.1 is proved.

We know from Propositions 3.4 and 3.5 that $s_1(r_{cr}) < s_3(r_{cr})$. Thus it is easy to see that $r = r_{cr}$ is a singular point with respect to $r$ in the asymptotic estimate of $E(N^2)$.

## 6. Conclusions and future work

In this paper we introduced the concept of *average similarity degree* to characterize how the solutions of random *k*-SAT and random CSPs are similar to each other. The main conclusion is that under certain conditions, as *r* (the ratio of constraints to variables) increases, the limit of average similarity degree when the number of variables approaches infinity exhibits phase transitions at a threshold point, shifting from a smaller value to a larger value abruptly. It should be mentioned that we can also define the *distance* between the two assignments in an assignment pair as a measure of how they are different from each other, i.e. the ratio of the number of variable where the two assignments differ to the total number of variables. Following the definition of average similarity degree, the *average distance* between solutions, denoted by $d_{av}$, can also be defined. It is easy to verify that the average similarity and the average distance satisfy the following equation

$$s_{av} + d_{av} = 1. \tag{6.1}$$

We can therefore immediately reach a conclusion from Theorems 1.1 and 1.2 that under the same conditions, the average distance will also exhibit phase transitions at the threshold point $r = r_{cr}$, shifting from a larger value to a smaller value abruptly. This conclusion implies that the solution space will suddenly shrink at the threshold point $r = r_{cr}$. Numerical calculations show that for random 5-SAT and random 6-SAT, the values of $r_{cr}$ are approximately 21.6 and 42.9 respectively. As a comparison, it was shown empirically that for random 5-SAT and random 6-SAT, the phase transition in solubility occurs when *r* is approximately 21.9 and 43.2 respectively[10]. So it would be interesting to investigate either theoretically or experimentally the relation between the phase transition in average similarity degree and the phase transition in solubility.

Our results suggest that the solutions of a random CSP instance will abruptly condense into a much smaller space when *r* crosses the phase transition point in average similarity degree. What can we learn from this study? Intuitively, if the solutions of a random instance are distributed in a smaller space, it might make search algorithms use more time to find a solution in the space of assignments and so harder to determine if this instance is soluble. As shown above, for random 5-SAT and random 6-SAT, the phase transition points in average similarity degree are very close to the corresponding phase transition points in solubility. Therefore, we can say that the finding of the phase transition in average similarity degree provides some new insights into understanding why there is a sharp increase in the hardness of solving random instances near the phase transition point in solubility. One possible application of it might be in the design of search algorithms. For example, when we solve a random CSP instance, it might be useful to determine the location of this instance before the search procedure begins. If it is located in the area where $r > r_{cr}$, then we can improve the efficiency of algorithms by using some information, e.g. the value of average similarity degree, to efficiently identify the location and structure of the solution space.

**Acknowledgement**

We wish to thank an anonymous referee and the guest editor Jianer Chen for helpful comments and suggestions which greatly improved the presentation of this paper.